\documentclass[10pt,twocolumn,letterpaper]{article}

\usepackage{times}
\usepackage{epsfig}
\usepackage{amssymb}
\usepackage{multirow}
\usepackage{amsmath}
\usepackage{booktabs}
\usepackage[T1]{fontenc}
\usepackage{bbm}
\usepackage{bbding}
\usepackage{colortbl}
\usepackage{threeparttable}
\usepackage{makecell}
\usepackage{pifont}
\usepackage{subcaption}
\usepackage{listings}
\usepackage{caption}
\usepackage[pagenumbers]{iccv} 

\newcommand{\keypoint}[1]{\vspace{0.1cm}\noindent\textbf{#1}}

\definecolor{iccvblue}{rgb}{0.21,0.49,0.74}
\usepackage[pagebackref,breaklinks,colorlinks,allcolors=iccvblue]{hyperref}

\def\paperID{*****} 
\def\confName{ICCV}
\def\confYear{2025}

\title{Learning Generalizable Prompt for CLIP with Class Similarity Knowledge}

\author{Sehun Jung \quad \quad Hyang-won Lee \\
Department of Computer Science and Engineering, Konkuk University\\
{\tt\small \{qhsl1213, leehw\}@konkuk.ac.kr}
}

\begin{document}
\maketitle
\begin{abstract}
In vision-language models (VLMs), prompt tuning has shown its effectiveness in adapting models to downstream tasks. However, learned prompts struggle to generalize to unseen classes, as they tend to overfit to the classes that are targeted during prompt tuning. Examining failure cases, we observed that learned prompts disrupt the semantics of unseen classes, generating text embeddings with incorrect semantic relationships among classes. To address this, we propose \textbf{S}imilarity \textbf{A}lignment \textbf{R}egularization (\textbf{SAR}), which regularizes learnable prompts to preserve the semantic relationships among classes captured by hand-crafted prompts. Specifically, we first obtain novel classes related to base classes using ChatGPT-4o and utilize them as potential unseen classes during prompt tuning. Then, by targeting both base and novel classes, SAR aligns the similarity relationships among text embeddings generated by learnable prompts with the similarity relationships from hand-crafted prompts. Extensive experiments applying SAR to existing prompt tuning methods demonstrate its effectiveness in improving generalization to unseen classes.
\end{abstract}    
\section{Introduction}
\label{sec:intro}

Recently, pre-trained foundation models that can be utilized for various downstream tasks have emerged across diverse fields. In the vision-language domain, CLIP \cite{radford2021clip} is recognized as one of the representative models. Trained on 400 million image-text pairs from the web, CLIP is able to capture rich semantic relationships between visual and textual information. This capability enables CLIP to achieve competitive zero-shot evaluation performance across a wide range of downstream tasks. One of the key findings in \cite{radford2021clip} is that using task-specific prompts can improve zero-shot performance by providing the model with additional context. For example, in EuroSAT dataset, which features satellite images depicting various types of land cover, using the prompt ``\texttt{a satellite photo of a}'' helps the model better understand the task, resulting in a notable enhancement in classification accuracy.

Instead of manually designing prompts, CoOp \cite{zhou2022learning} proposes a prompt tuning approach that learns task-specific prompts. In this method, learnable vectors in the word embedding space are trained using image-text pairs from target tasks and serve as prompts. This approach achieves significant performance improvements over using hand-crafted prompts, emphasizing the importance of prompt designing. However, learned prompts often fail to generalize to unseen classes that were not targeted during prompt tuning, resulting in substantial performance degradation. Extensive research has focused on learning generalizable prompts by exploring both architectural improvements \cite{zhou2022conditional, khattak2023maple, lee2023read, zhang2024dept} and regularization methods \cite{zhu2023prompt, yao2023kgcoop, cho2023distribution, ding2024lobg}. A widely adopted regularization method minimizes the distance between the text embeddings generated by learnable prompts and those generated by hand-crafted prompts during prompt tuning \cite{yao2023kgcoop, khattak2023promptsrc, roy2024coprompt, yao2024tcp, ding2024lobg}. This approach improves the generalization ability of learnable prompts by preserving the general knowledge captured by hand-crafted prompts. However, a significant limitation of this method lies in its focus on individual classes in isolation, without considering the relationships among classes. As a result, inter-class consistency is not explicitly accounted for during prompt tuning, which indicates that there is room for further improvement in generalization.

In this work, we first investigate how prompts learned from base classes fail to generalize to unseen classes in terms of semantic disruptions. We observe that when the text embeddings generated by learned prompts form incorrect semantic relationships with other classes, they act as low-quality classifiers, leading to poor generalization. In contrast, hand-crafted prompts capture meaningful semantic relationships among classes due to their strong generalization ability. Motivated by these findings, we propose \textbf{S}imilarity \textbf{A}lignment \textbf{R}egularization (\textbf{SAR}), a method for regularizing learnable prompts to preserve the semantic relationships captured by hand-crafted prompts. Specifically, we first obtain novel classes semantically aligned with the base classes by utilizing ChatGPT-4o. For both base and novel classes, SAR aligns the similarity relationships among text embeddings generated by learned prompts with those relationships from hand-crafted prompts. By preserving the meaningful semantic relationships among classes, SAR enables learnable prompts to effectively capture the semantics of unseen classes. Additionally, at each parameter update step, regularization is applied to the similarities computed on randomly sampled classes, instead of using all classes. This approach mitigates overfitting by introducing noise during prompt tuning and further improves the generalization of learned prompts to unseen classes.

As a regularizer, SAR can be applied to existing prompt tuning models without any architectural modifications. We validate the effectiveness of SAR on various baselines, including textual prompt tuning models: CoOp \cite{zhou2022learning}, KgCoOp \cite{yao2023kgcoop}, TCP \cite{yao2024tcp} and multi-modal prompt tuning models: MaPLe \cite{khattak2023maple}, CoPrompt \cite{roy2024coprompt}. Notably, in base-to-new generalization experiments, SAR consistently improves overall accuracy on new classes across 11 datasets for five baselines, while either preserving or enhancing the base accuracy. Our contributions are summarized as follows:
\begin{itemize}[itemsep=0.5em, topsep=0.5em]
    \item We experimentally analyze how learned prompts fail to generalize to unseen classes, focusing on semantic disruptions.
    \item We propose SAR, a method to guide learnable prompts to capture meaningful semantic relationships among classes. SAR utilizes novel classes generated by ChatGPT-4o as potential unseen classes during prompt tuning.
    \item We incorporate a random embedding sampling strategy into SAR, to mitigate overfitting.
    \item We demonstrate the effectiveness of SAR in improving generalization to unseen classes through extensive experiments across 11 datasets and five baselines.
\end{itemize}

\section{Related Work}
\label{sec:relatedworks}

\keypoint{Pre-trained Vision-Language Models.}
Large-scale pre-training is essential for VLMs to develop a comprehensive understanding of the relationships between images and text. The effectiveness of pre-trained VLMs on downstream tasks largely depends on the design of pre-training tasks, which define what the model learns from the data \cite{du2022survey}. Among these tasks, image-text matching (ITM) and masked language modeling (MLM) are widely adopted due to their complementary roles \cite{li2019visualbert, kim2021vilt, li2021align, bao2022vlmo}. ITM enables models to focus on the global semantics of images and text, facilitating coarse-grained alignments. In contrast, MLM encourages models to extract information from objects within images to predict masked language tokens, thereby promoting fine-grained alignments. Meanwhile, transformers \cite{vaswani2017attention} play a key role in VLMs by serving as powerful contextualizers, enabling VLMs to model complex relationships between modalities. Among VLMs, CLIP \cite{radford2021clip} employs a dual-encoder architecture trained with contrastive learning. Despite being trained on noisy web data, it demonstrates effectiveness on various downstream tasks, including complex tasks such as monocular depth estimation \cite{auty2023learning}, visual question answering \cite{song2022clip}, and instance segmentation \cite{ding2022open}.

\keypoint{Fine-tuning VLMs for Downstream Tasks.} Fine-tuning all parameters in pre-trained VLMs on limited data of downstream tasks is prone to losing the rich representations learned by the model and overfitting. To address this issue, parameter-efficient fine-tuning methods have emerged as an alternative to conventional fine-tuning approaches. One such method, CLIP-Adapter \cite{gao2024clip}, trains small adapters that transform the output embeddings of CLIP encoder into task-useful features. Another approach is prompt tuning, which focuses on tailoring the model’s input. In CoOp \cite{zhou2022learning}, a set of learnable vectors is added to the input embeddings to guide the encoder to generate task-useful embeddings. CoCoOp \cite{zhou2022conditional} conditioned the prompts on image features, generating image-adaptive prompts that improve generalization to unseen classes. MaPLe \cite{khattak2023maple} and PromptSRC \cite{khattak2023promptsrc} extend adaptability to downstream tasks by adding prompts to both text and image inputs. In contrast to the above approaches, some fine-tuning methods focus on updating only specific parameters, such as bias and normalization terms \cite{song2022clip}.

\begin{figure*}[t]
    \setlength{\abovecaptionskip}{0.1cm}  
    \setlength{\belowcaptionskip}{-0.3cm} 
    \centering
    \includegraphics[width=1\linewidth]{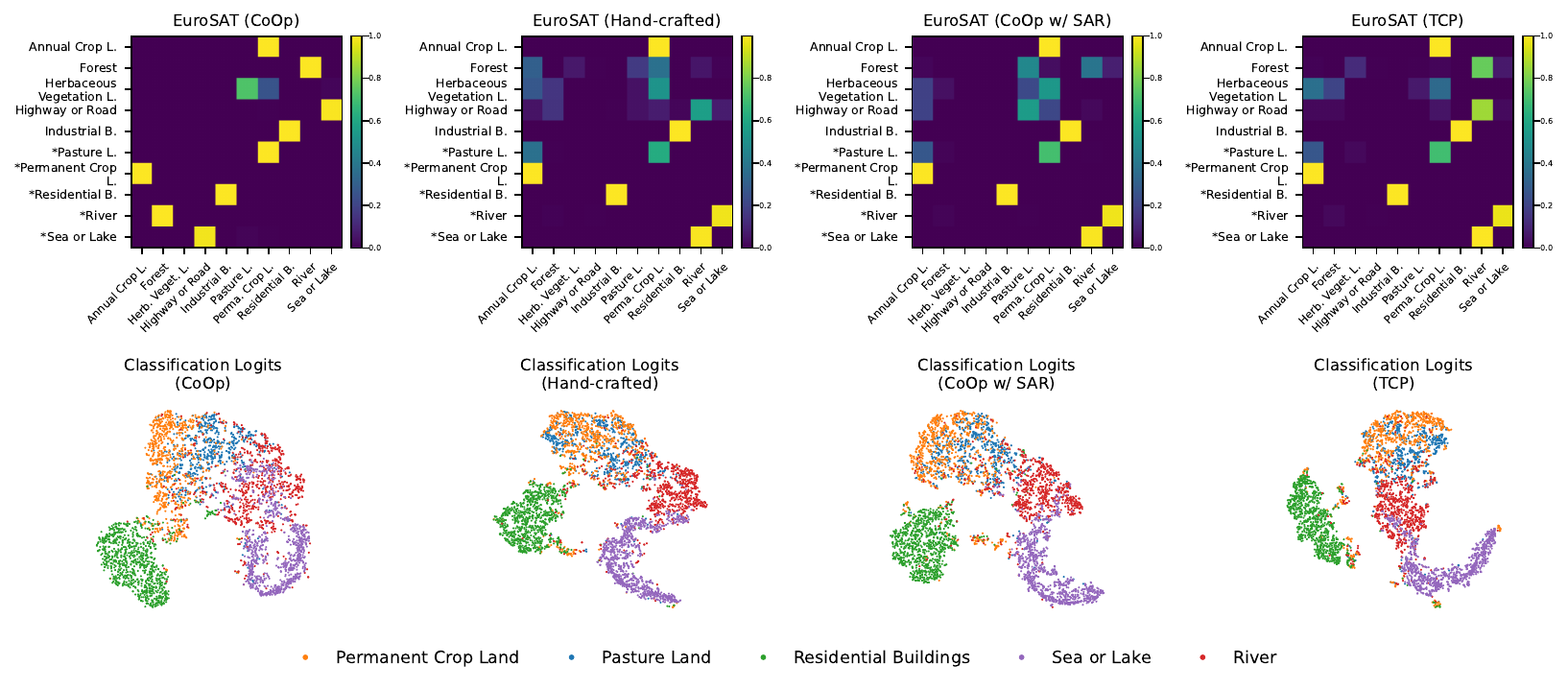} 
    \caption{Prompt generalization evaluation. \textbf{(Top)} Heatmap visualization of similarity distribution matrices computed over all (base+new) classes. From left to right: 1) $\mathbf{P}_{\mathtt{CoOp}}$, 2) $\mathbf{P}_{\mathtt{hand}}$, 3) produced by prompts learned by CoOp with SAR applied, and 4) produced by prompts learned by TCP \cite{yao2024tcp}. In class names, \textup{L.} and \textup{B.} are abbreviations of $Land$ and $Building$, respectively. An asterisk (*) before a class name indicates that it is a new class, which was not used during prompt training. \textbf{(Bottom)} t-SNE scatterplots of logits for test images from new classes. In CoOp, the logits points corresponding to images of $River$ and $Sea\ or\ Lake$ are broadly distributed, forming an ambiguous cluster boundary. In contrast, such issues are not observed in the logits visualization of CoOp with SAR applied, thank to the guiding of SAR.}
    \label{fig:motivation}
\end{figure*}

\keypoint{Regularization for Prompt Tuning.} Various regularization techniques have been explored to learn generalizable prompts. ProGrad \cite{zhu2023prompt} ensures that the gradient for prompt tuning does not conflict with the gradient used to preserve the general knowledge of CLIP. Specifically, if the angle between the two gradient vectors is obtuse, the gradient for prompt tuning is projected to be orthogonal to the other gradient, and is used for the update. KgCoOp \cite{yao2023kgcoop} minimizes the distance between text embeddings generated by learnable prompts and those generated by hand-crafted prompts, preserving the general knowledge captured by hand-crafted prompts. LASP \cite{bulat2023lasp} introduces a text-to-text cross-entropy loss to ensure that text embeddings generated by learnable prompts are correctly classified as those generated by hand-crafted prompts for the same class. DAPT \cite{cho2023distribution} enforces a uniform distribution of text embeddings on a hypersphere to minimize overlap while encouraging image embeddings of the same class to be positioned closer together, to achieve better alignment. TPR \cite{chentpr} maximize Pearson correlation coefficient computed between the pairwise cosine similarity matrices of the original CLIP text embeddings and the learned text embeddings, to preserve the class topology. In this process, they utilize the textual descriptions of both base and new classes in the dataset. LOBG \cite{ding2024lobg} preserves the structural topology of image embeddings captured by hand-crafted prompts, which is achieved by maintaining local angular relationships among image embeddings. Unlike these methods, our method SAR utilize the similarity distribution among text embeddings, to effectively capture the relational similarities among classes.
\section{Prompt Tuning with SAR}
In this section, we present the implementation of SAR and its integration within the prompt tuning framework.

\subsection{Preliminaries}
\keypoint{Contrastive Language-Image Pre-training (CLIP) \cite{radford2021clip}.} CLIP is a vision-language model pre-trained on 400 million image-text pairs, designed to align semantic relationships between image and text modalities. It consists of an image encoder $\theta(\cdot)$ and a text encoder $\phi(\cdot)$ that map their respective inputs into a shared embedding space. CLIP is trained using contrastive learning, where embeddings of matched image-text pairs are pulled closer together, while those of unmatched pairs are pushed apart. We can perform zero-shot image classification using CLIP by comparing the matching score, which is cosine similarity between the image embedding and the text embedding. Using appropriate prompts can significantly enhance CLIP’s classification performance. Finding task-optimized prompts is referred to as prompt engineering, which often relies on a trial-and-error process.

\keypoint{Context Optimization.} CoOp \cite{zhou2022learning} proposes a framework that trains prompts via a classification task using image-text pairs of target tasks. In CoOp, $P$ trainable vectors $[\boldsymbol{v}_1, \boldsymbol{v}_2, \dots, \boldsymbol{v}_P]$ serve as prompts. These vectors are added to $\boldsymbol{c}_i$, the word embedding(s) of the $i$-th class name $\boldsymbol{w}_i$, and passed to the text encoder to compute the text embedding $\boldsymbol{g}_i = \phi([\boldsymbol{v}_1, \boldsymbol{v}_2, \dots, \boldsymbol{v}_P, \boldsymbol{c}_i])$. Similarly, the image embedding $\boldsymbol{f}$ is computed by passing each image $x$ through the image encoder $\theta(\cdot)$. The probability that $\boldsymbol{x}$ is predicted as $\boldsymbol{w}_i$ is computed as:

\begin{equation}
p(\boldsymbol{w}_i|\boldsymbol{x}) = \frac{\exp(\cos(\boldsymbol{g}_i,\boldsymbol{f}) / \tau)}{\sum_{j=1}^C \exp(\cos(\boldsymbol{g}_j,\boldsymbol{f}) / \tau)},
\end{equation}
where $\cos(\cdot, \cdot)$ denotes cosine similarity, $C$ is the number of classes, and $\tau$ is the learned temperature parameter of CLIP. To optimize prompts for target tasks, trainable vectors are updated by cross-entropy loss:

\begin{equation}
\mathcal{L}_{\text{ce}}=-\sum_{i} \boldsymbol{y}_i \log p(\boldsymbol{w}_i|\boldsymbol{x}),
\end{equation}
where $\boldsymbol{y}$ is the one-hot encoded label. During prompt tuning, all weights in the image and text encoders of CLIP are fixed.

\begin{figure*}[ht]
    \setlength{\abovecaptionskip}{0.2cm}  
    \setlength{\belowcaptionskip}{-0.4cm} 
    \centering
    \includegraphics[width=1.0\linewidth]{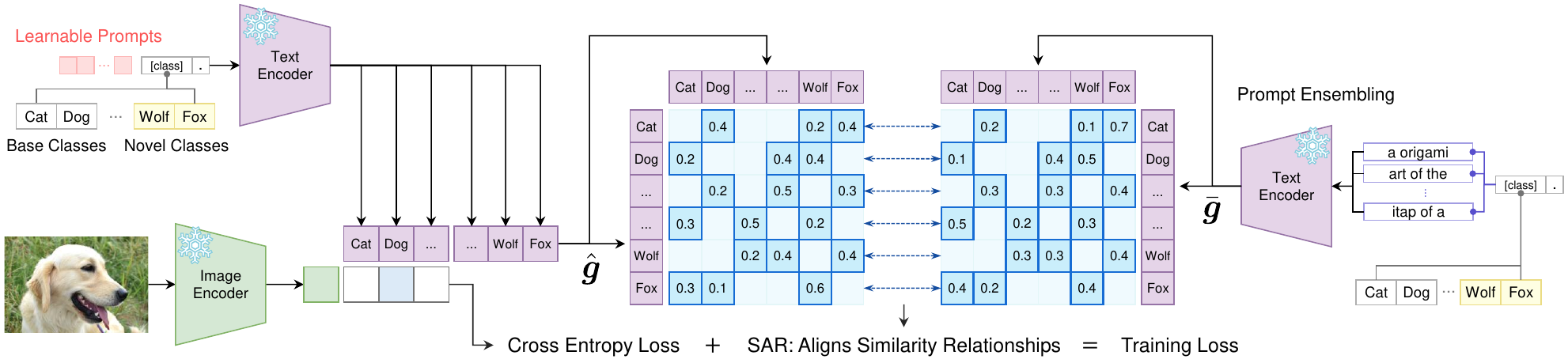} 
    \caption{Overview of how Similarity-Alignment Regularization (SAR) operates in prompt tuning. SAR targets the base and novel classes, aligning the semantic relationships among text embeddings generated by learnable prompts with those relationships from ensembled hand-crafted prompts. Specifically, this alignment is achieved by minimizing the KL divergence between the corresponding similarity distributions. To mitigate overfitting, random embedding sampling is employed instead of computing similarities across all classes.}
    \label{fig:framework}
\end{figure*}

\subsection{Semantic Disruption by Learned Prompts}
Now we analyze how the semantically misaligned text embeddings, generated by learned prompts, lead to poor generalization.

\keypoint{Computing Semantic Similarities among Classes.} We begin by describing how to compute a similarity distribution matrix, which shows the semantic relationships among text embeddings. Let $\{\boldsymbol{g}_j\}_{j=1}^{N}$ denote a set of text embeddings corresponding to $N$ arbitrary classes, where $\boldsymbol{g}_j \in \mathbb{R}^{d}\ \forall j$. Since matched image-text pairs are located closer in CLIP’s embedding space, text pairs with similar semantics also tend to be closer, allowing us to measure their semantic similarities via cosine similarity. We can compute the cosine similarity matrix $\mathbf{C} \in \mathbb{R}^{N \times N}$ with entries $\mathbf{C}_{ij}=\cos(\boldsymbol{g}_i, \boldsymbol{g}_j)$. To highlight relational similarities among classes, we compute the similarity distribution matrix $\mathbf{P}$, by applying a softmax function to each row of $\mathbf{C}$, excluding its diagonal elements (which represent self-similarity) from the computation:
\begin{equation}
\mathbf{P}_{ij} =
\begin{cases} 
0, & \text{if } i = j, \\
\frac{\exp\left( \cos(\boldsymbol{g}_i, \boldsymbol{g}_j) / \tau \right)}{\sum_{k \neq i} \exp\left( \cos(\boldsymbol{g}_i, \boldsymbol{g}_k) / \tau \right)}, & \text{if } i \neq j.
\end{cases}
\end{equation}
Note that the matrix $\mathbf{P}$ is not necessarily symmetric, as the normalization can differ for each row.

\keypoint{Prompt Generalization Evaluation.} To evaluate the generalization of learned prompts to unseen classes, we perform base-to-new generalization experiments. Specifically, we first train prompts using image-text pairs from only half of dataset's classes (base classes). Then, we conduct zero-shot evaluation on the test data of remaining half classes (new classes) using the learned prompts. During this process, we compare the two similarity distribution matrices computed over all (base+new) classes: 1) $\mathbf{P}_{\mathtt{CoOp}}$, produced by learned prompts, and 2) $\mathbf{P}_{\mathtt{hand}}$, produced by hand-crafted prompts. Additionally, we examine the logit patterns for the test images. Specifically, we transform the logit vector for each image into a 2-dimensional vector using t-SNE \cite{van2008visualizing} and visualize the results.

The experimental results on EuroSAT \cite{helber2019eurosat} are presented in \Cref{fig:motivation}. The leftmost heatmap, corresponding to $\mathbf{P}_{\mathtt{CoOp}}$, reveals that the learned prompts generate semantically misaligned text embeddings for certain new classes. For example, in $\mathbf{P}_{\mathtt{CoOp}}$, the class $River$ shows a higher similarity to $Forest$ than to $Sea\ or\ Lake$, despite the latter being semantically closer. Similarly, the similarity between $Sea\ or\ Lake$ and $Highway\ or\ Road$ is abnormally high, even though they are semantically unrelated. Moreover, the learned prompts fail to preserve subtle semantic distinctions among classes, forming overly confident associations. In contrast, $\mathbf{P}_{\mathtt{hand}}$ captures the more meaningful semantic relationships and better reflects subtle distinctions among classes. The logits scatterplots at the bottom further illustrate that semantically misaligned text embeddings lead to poor generalization in the classification task. In the scatterplot for CoOp, the logits for $Sea\ or\ Lake$ and $River$ form poorly defined clusters that overlap significantly, showing that the text embeddings struggle to capture the visual concepts of the two classes. In contrast, in the scatterplot for the hand-crafted prompt, the clusters for the two classes are clearly separated, highlighting its superior capability in generalization.

\subsection{Similarity Alignment Regularization}
To prevent semantic disruptions caused by learned prompts and improve generalization, we propose Similarity Alignment Regularization (SAR). \Cref{fig:framework} illustrates the operation of SAR in prompt learning.

\keypoint{Prompt Learning with Novel Classes.} We use ChatGPT-4o to generate novel classes semantically aligned with the base classes, utilizing them as potential unseen classes during prompt tuning. Specifically, we provide ChatGPT-4o with a list of the base classes and instruct it to generate semantically aligned novel classes. The prompt given to ChatGPT-4o is provided in \underline{\textbf{Supp. Material. A}}. To obtain more robust and representative embeddings for each class, we use ensembled text embeddings from multiple hand-crafted prompts. Let $T$ be the number of hand-crafted prompts used for prompt ensembling. For $j$-th class, the ensembled text embedding is computed as $\bar{\boldsymbol{g}}_j=\frac{1}{T}\sum_{i=1}^T \boldsymbol{g}_j^i$, where $\boldsymbol{g}_j^i$ is the text embedding of $j$-th class generated by the $i$-th prompt. The hand-crafted prompts used for ensembling are listed in \underline{\textbf{Supp. Material. B}}. Let $\mathcal{G}_{\mathtt{hand}}=\{\bar{\boldsymbol{g}}_j\}_{j=1}^{M}$ denote the set of text embeddings for both base and novel classes, generated by the hand-crafted prompts. In contrast, $\mathcal{G}_{\mathtt{CoOp}}=\{\hat{\boldsymbol{g}}_j\}_{j=1}^{M}$ represents the set of text embeddings generated by the learnable prompts.

For learnable prompts to preserve meaningful semantic relationships among classes, SAR aim to align the semantic relationships in $\mathcal{G}_{\mathtt{CoOp}}$ with those in $\mathcal{G}_{\mathtt{hand}}$ during prompt tuning.
A straightforward way to achieve this is to construct similarity distribution matrices for $\mathcal{G}_{\mathtt{CoOp}}$ and $\mathcal{G}_{\mathtt{hand}}$, respectively, and minimize the KL divergence \cite{kullback1951information} between them during prompt learning. This method simultaneously takes into account all pairwise semantic similarities between the text embeddings. To mitigate overfitting, however, we introduce randomness into the regularization process through random embedding sampling.

\keypoint{Computing Similarities for Sampled Classes.} For each $\hat{\boldsymbol{g}}_j$, we compute similarities with $K$ other embeddings that are randomly sampled, rather than considering all other embeddings. Let $\mathcal{I}_j$ denote the set of $K$ indices of the sampled embeddings, ensuring that $j \notin \mathcal{I}_j$ to exclude self-similarity. Now the similarity disbribution matrix $\mathbf{P}_{\mathtt{CoOp}}^{\mathcal{I}}$ is computed as:
\renewcommand{\arraystretch}{1.5}
\begin{equation}
\mathbf{P}_{\mathtt{CoOp}}^{\mathcal{I}} = 
\begin{bmatrix}
\sigma\Big(\big[\cos(\hat{\boldsymbol{g}}_1, \hat{\boldsymbol{g}}_k) \,|\, k \in \mathcal{I}_1\big]\Big) \\
\sigma\Big(\big[\cos(\hat{\boldsymbol{g}}_2, \hat{\boldsymbol{g}}_k) \,|\, k \in \mathcal{I}_2\big]\Big) \\
\vdots \\
\sigma\Big(\big[\cos(\hat{\boldsymbol{g}}_M, \hat{\boldsymbol{g}}_k) \,|\, k \in \mathcal{I}_M\big]\Big)
\end{bmatrix}
\in \mathbb{R}^{M \times K},
\end{equation}
where $\sigma(\cdot)$ denotes the softmax function with the CLIP-learned temperature $\tau$. Using $\mathcal{G}_{\mathtt{hand}}$, we also compute $\mathbf{P}_{\mathtt{hand}}^{\mathcal{I}} \in \mathbb{R}^{M \times K}$ with the same family $\mathcal{I}$ of index sets, as the target of $\mathbf{P}_{\mathtt{CoOp}}^{\mathcal{I}}$. This approach allows learnable prompts to capture the relative semantic similarities among classes more effectively.

\keypoint{Randomized Similarity Alignment.} SAR minimizes the KL divergence between the corresponding rows of $\mathbf{P}^{\mathcal{I}}_{\mathtt{CoOp}}$ and $\mathbf{P}^{\mathcal{I}}_{\mathtt{hand}}$:

\begin{equation}
\mathcal{L}_{\text{SAR}} = \frac{1}{M} \sum_{i=1}^M \mathcal{D}_{\mathcal{KL}}(\mathbf{P}^{\mathcal{I}}_{\mathtt{CoOp},i} \parallel \mathbf{P}^{\mathcal{I}}_{\mathtt{hand},i}).
\end{equation}
where $\mathcal{I}$ is newly sampled at every single parameter update step during prompt tuning.
Finally, the training objective for SAR-applied prompt tuning is:

\begin{equation}
\mathcal{L}=\mathcal{L}_{\text{ce}}+\lambda \mathcal{L}_{\text{SAR}}
\end{equation}
where $\lambda$ is the hyperparameter that controls the strength of the regularization.
\section{Experiments}
We evaluate the effectiveness of SAR in 1) base-to-new generalization and 2) domain generalization. To demonstrate its compatibility with existing prompt tuning models, SAR is applied to textual prompt tuning models: CoOp \cite{zhou2022learning}, KgCoOp \cite{yao2023kgcoop}, TCP \cite{yao2024tcp}, and multi-modal prompt tuning models: MaPLe \cite{khattak2023maple}, CoPrompt \cite{roy2024coprompt}.

\keypoint{Datasets.}
For base-to-new generalization, we evaluate our method on 11 datasets. ImageNet \cite{deng2009imagenet} and Caltech-101 \cite{fei2004learning} are used for generic object classification, while fine-grained classification is conducted on OxfordPets \cite{parkhi2012cats}, StanfordCars \cite{krause20133d}, Flowers102 \cite{nilsback2008automated}, Food-101 \cite{bossard2014food}, and FGVCAircraft \cite{maji2013fine}. For domain-specific classification tasks, we use EuroSAT \cite{helber2019eurosat} for satellite image classification, UCF101 \cite{soomro2012ucf101} for action recognition, DTD \cite{cimpoi2014describing} for texture classification, and SUN397 \cite{xiao2010sun} for scene recognition. For domain generalization, ImageNet serves as the source dataset, while ImageNet-V2 \cite{recht2019imagenet}, ImageNet-Sketch \cite{wang2019learning}, ImageNet-A \cite{gao2022generating}, and ImageNet-R \cite{hendrycks2021many} act as target domains.

\begin{table}[!t]
    \small \centering
    \renewcommand{\arraystretch}{0.9}
    \setlength{\tabcolsep}{8pt}
    \scalebox{0.95}[0.95]{
        \begin{tabular}{l cc | c }
        \toprule
        Method  & Base & New & H \\
        \midrule
        CoOp & 82.35 & 66.61 & 73.65 \\
        \hspace{0.0em}+ $\mathcal{L}_{\text{SAR}}$ & 82.46 & 71.91 & 76.82 \\
        \hspace{0.0em}+ Prompt Ensembling & \textbf{82.82} & 73.39 & 77.82 \\
        \rowcolor{gray!20}
        \hspace{0.0em}+ Random Embedding Sampling & 82.79 & \textbf{73.75} & \textbf{78.01} \\
        \bottomrule
        \end{tabular}
    }
    \caption{Evaluation of each component's contribution to SAR. Results are averaged across 11 datasets. H refers to harmonic mean.}
    \label{tab:component_ab}
\end{table}

\keypoint{Implementation details.}
All experiments are conducted with the CLIP ViT-B/16 model. For each dataset, 200 novel classes generated by ChatGPT-4o are used for SAR. The embedding sampling size $K$ is fixed at 64 across all experiments. In base-to-new generalization, for CoOp and KgCoOp, the number of training epochs is set to 100 by default to reduce training time. Except for this, we follow the same training settings, such as learning rate, epochs, and training schedules, as the baseline models. The hyperparameter $\lambda$ for SAR is adjusted based on the difficulty of optimizing its loss and the characteristics of the dataset. For TCP, $\lambda$ is set to 0.1 across all datasets. On EuroSAT, $\lambda$ is set to 0.5 in CoOp and 0.75 in both MaPLe and CoPrompt. On DTD and UCF101, $\lambda$ is set to 0.5 in all three models. On FGVCAircraft, $\lambda$ is set to 0.5 in CoOp, 0.75 in MaPLe, and 0.8 in CoPrompt. For all remaining datasets, $\lambda$ is set to 0.1 in all three models. The final performance is reported as the average over seeds 1, 2, and 3 to ensure a fair evaluation. We reproduce all baseline performances using the original code provided by the authors.

\subsection{Ablation Study}  
\keypoint{Stepwise Analysis of SAR's Effectiveness.}
To analyze the contribution of each component in SAR, we conduct a base-to-new generalization experiment in the 16-shot setting, using CoOp as the baseline. The results, averaged across 11 datasets, are presented in \Cref{tab:component_ab}. First, CoOp exhibits a significant gap between base and new accuracies, revealing its limited ability to generalize to unseen classes. By introducing $\mathcal{L}_{\text{SAR}}$, this limitation is effectively addressed, leading to a substantial improvement in new accuracy (+5.3\%) while also slightly enhancing base class accuracy (+0.11\%). Building on this, the integration of prompt ensembling further enhances both base and new accuracies by guiding the learnable prompts with more robust semantic relationships among classes. Lastly, incorporating random embedding sampling further boosts new accuracy with a negligible decrease in base accuracy. Overall, SAR improves CoOp's base accuracy from 82.35\% to 82.79\% (+0.44\%), new accuracy from 66.61\% to 73.75\% (+7.14\%), and the harmonic mean (H) from 73.65\% to 78.01\% (+4.36\%), demonstrating its effectiveness by improving all three metrics.

\begin{figure}[!t]
    \setlength{\abovecaptionskip}{1pt}
    \centering
    \includegraphics[width=1.0\linewidth]{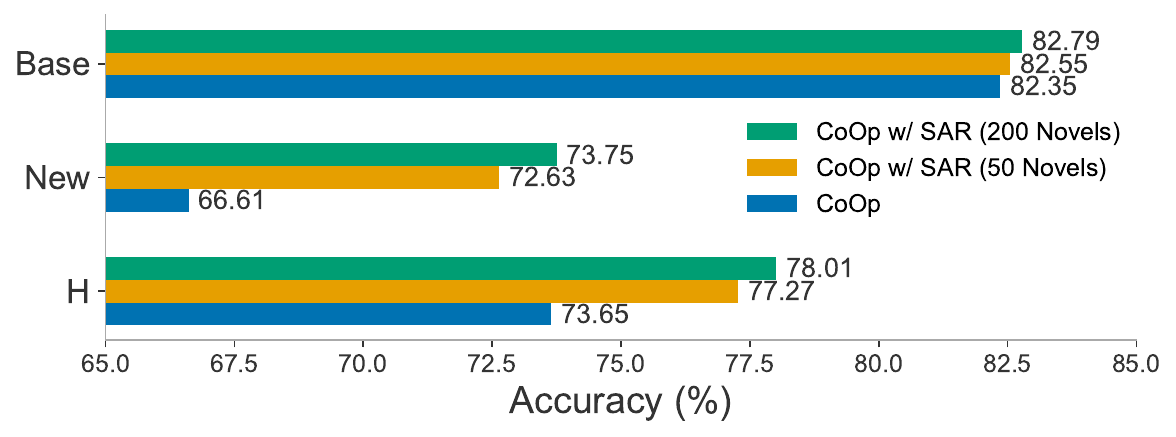}
    \caption{Effect of number of novel classes on performance in 16-shot setting. The results are averaged across 11 datasets.}
    \label{fig:novelnum_ab}
\end{figure}

\begin{figure}[!t]
    \centering
    \includegraphics[width=1.0\linewidth]{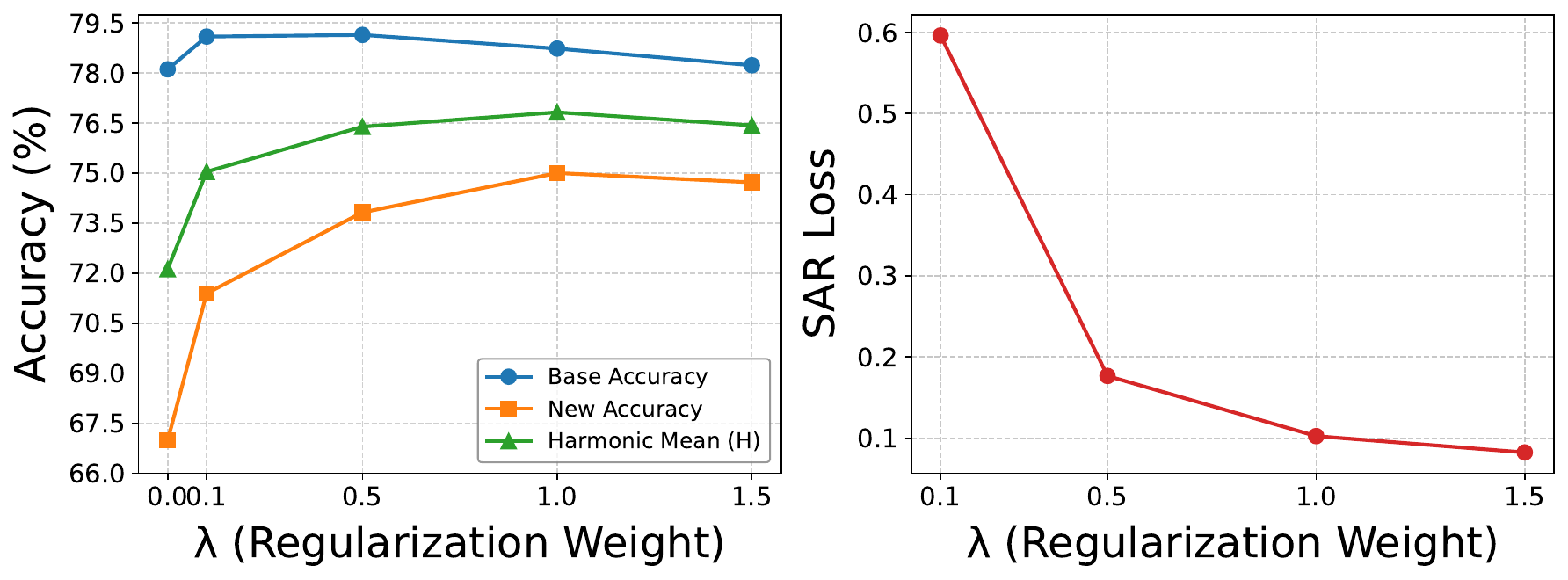}
    \caption{\textbf{Left}: Effect of regularization weight $\lambda$ on performance in 4-shot setting. \textbf{Right}: Trend of SAR loss as $\lambda$ increases. The results are averaged across 11 datasets.}
    \label{fig:weight_ab} 
\end{figure}

\keypoint{Impact of Number of Novel Classes on SAR.}
We evaluate the impact of the number of novel classes on the effectiveness of SAR. To this end, we train prompts by applying SAR under two different numbers of novel classes: one using 50 novel classes and the other using 200 novel classes. In the 50-class setup, the classes are randomly selected from the set of 200 novel classes, and the embedding sampling size is fixed at 16. As the results in \Cref{fig:novelnum_ab} show, using a larger number of novel classes leads to greater performance improvements. With more novel classes, richer similarity information among semantically diverse classes can be obtained, enabling a more refined and accurate capture of the semantics of unseen classes. Notably, even when using only 50 novel classes for SAR, significant performance improvements over the baseline are observed.

\keypoint{Impact of Regularization Weight $\lambda$ on SAR.}
We evaluate the impact of the regularization weight $\lambda$ on performance. The experimental results under 4-shot setting are presented in \Cref{fig:weight_ab}. The left plot in the figure shows the variations in base accuracy, new accuracy, and their harmonic mean (H) as $\lambda$ changes. Increasing $\lambda$ up to 1.0 leads to a significant improvement in new accuracy, showing a clear association between SAR's mechanism and the enhancement of generalization performance to new classes. However, when $\lambda$ becomes excessively large (\eg 1.5), all accuracies slightly decline due to the effects of overly strong regularization. The right plot in \Cref{fig:weight_ab} depicts the average SAR loss at the final epoch of the model, averaged across all datasets. As $\lambda$ increases, SAR loss decreases, which corresponds to the improvement in new accuracy observed in the left plot.

\begin{table}[!t]
    \small \centering
    \renewcommand{\arraystretch}{1.0}
    \setlength{\tabcolsep}{4pt}
    \resizebox{\columnwidth}{!}{
    \begin{tabular}{lcccc | c}
    \toprule
        Method & {Params.} & {Train time} & {Infer time} & {Memory} & {H} \\ 
        \midrule
        CoOp & 2048 & 39m 47s & 5.57ms & 11.36GiB & 73.65 \\
        \rowcolor{gray!20}
        \textbf{+SAR (50)} & \textbf{+0} & \textbf{+8m 7s} & \textbf{+0ms} & \textbf{+0.86GiB} & \textbf{77.27} \textcolor{blue}{(+3.62\%)}\\
        \rowcolor{gray!20}
        \textbf{+SAR (200)} & \textbf{+0} & \textbf{+25m 37s} & \textbf{+0ms} & \textbf{+3.82GiB} & \textbf{78.01} \textcolor{blue}{(+4.36\%)}\\
        \bottomrule
            \end{tabular}
        } 
    \caption{Resource costs for introducing SAR on ImageNet using a single NVIDIA A100 GPU. The number in $(\cdot)$ is the number of novel classes used for SAR. H is averaged across 11 datasets.}
    \label{tab:incurred_resource}
\end{table}

\keypoint{Resource Costs for Introducing SAR.}
We report the additional resource requirements incurred by SAR on ImageNet in \Cref{tab:incurred_resource}. Since SAR does not modify the model architecture, it does not introduce any extra trainable parameters. Moreover, as SAR operates during training, it has no impact on inference time. However, implementing SAR requires computing text embeddings not only for the base classes but also for the novel classes. This additional computation in the text encoder, along with the computation of similarity among classes, increases memory usage and training time. Furthermore, additional training time increases with the number of update steps, as text embeddings are computed at every update. Among the 11 datasets, ImageNet has the largest training set, resulting in the highest number of update steps. Consequently, the additional training time is notably smaller for the other 10 datasets. The number of novel classes for SAR can be chosen by considering the tradeoff between resource cost and performance, as the additional memory usage and training time incurred by SAR grow with the number of novel classes.

\begin{table*}[htbp]
\setlength{\abovecaptionskip}{0.1cm}
\setlength{\belowcaptionskip}{-0.1cm}
\centering
\renewcommand{\arraystretch}{1.1}
\tabcolsep 0.12in
\footnotesize
\begin{tabular}{l|ccc|ccc|ccc|ccc}
  \hline
  \multicolumn{1}{l|}{\multirow{2}{*}{\makecell[c]{Method}}} & \multicolumn{3}{c|}{\textbf{Avg over 11 datasets}} & \multicolumn{3}{c|}{ImageNet} & \multicolumn{3}{c|}{Caltech101} & \multicolumn{3}{c}{OxfordPets} \\
  \cline{2-13}
  & Base & New & H & Base & New & H & Base & New & H & Base & New & H \\
  \hline
  CoOp \cite{zhou2022learning} & 82.35 & 66.61 & 73.65 & 76.26 & 67.56 & 71.65 & 98.09 & 92.21 & 95.06 & 93.75 & 95.70 & 94.71 \\
  \cellcolor{gray!20}{\textbf{+SAR}} &\cellcolor{gray!20}{\textbf{\textcolor{blue}{82.79}}} &\cellcolor{gray!20}{\textbf{\textcolor{blue}{73.75}}}  &\cellcolor{gray!20}{\textbf{\textcolor{blue}{78.01}}} & \textbf{76.61} & \textbf{69.95} & \textbf{73.13} & \textbf{98.09} & \textbf{94.91} & \textbf{96.47} & \textbf{94.45} & \textbf{96.25} & \textbf{95.34} \\
  \hline
  KgCoOp \cite{yao2023kgcoop} & 81.67 & 73.15 & 77.18 & 75.65 & 69.65 & 72.53 & 97.85 & 94.36 & 96.07 & 95.09 & 97.50 & 96.28 \\
  \cellcolor{gray!20}{\textbf{+SAR}} & \cellcolor{gray!20}{81.40} &\cellcolor{gray!20}{\textbf{\textcolor{blue}{73.98}}}  &\cellcolor{gray!20}{\textbf{\textcolor{blue}{77.51}}} & 75.63 & \textbf{69.88} & \textbf{72.64} & 97.74 & \textbf{94.47} & \textbf{96.08} & 95.07 & 96.70 & 95.88 \\
  \hline
  TCP \cite{yao2024tcp} & 83.89 & 75.21 & 79.31 & 77.10 & 69.66 & 73.19 & 98.13 & 94.80 & 96.44 & 94.51 & 97.07 & 95.77 \\
  \cellcolor{gray!20}{\textbf{+SAR}} & \cellcolor{gray!20}{83.78} &\cellcolor{gray!20}{\textbf{\textcolor{blue}{75.45}}}  &\cellcolor{gray!20}{\textbf{\textcolor{blue}{79.40}}} & 77.04 & \textbf{69.81} & \textbf{73.25} & \textbf{98.15} & 94.76 & 96.43 & 94.40 & 96.35 & 95.37 \\
  \hline
  MaPLe \cite{khattak2023maple} & 81.84 & 74.56 & 78.03 & 76.70 & 70.56 & 73.50 & 97.68 & 94.87 & 96.25 & 95.50 & 98.02 & 96.74 \\
  \cellcolor{gray!20}{\textbf{+SAR}} &\cellcolor{gray!20}{\textbf{\textcolor{blue}{82.25}}} &\cellcolor{gray!20}{\textbf{\textcolor{blue}{76.06}}}  &\cellcolor{gray!20}{\textbf{\textcolor{blue}{79.03}}} & \textbf{76.82} & \textbf{70.83} & \textbf{73.70} & \textbf{97.85} & 94.25 & 96.02 & 95.41 & 97.13 & 96.26 \\
  \hline
  CoPrompt \cite{roy2024coprompt} & 82.82 & 74.60 & 78.50 & 76.66 & 71.27 & 73.87 & 98.65 & 95.09 & 96.84 & 95.14 & 96.20 & 95.67 \\
  \cellcolor{gray!20}{\textbf{+SAR}} & \cellcolor{gray!20}{82.72} &\cellcolor{gray!20}{\textbf{\textcolor{blue}{76.52}}}  &\cellcolor{gray!20}{\textbf{\textcolor{blue}{79.50}}} & 76.55 & 71.18 & 73.77 & \textbf{98.69} & 94.87 & 96.74 & 95.09 & \textbf{96.44} & \textbf{95.76} \\
  \hline
    \multicolumn{1}{l|}{\multirow{2}{*}{\makecell[c]{{Method}}}}  & \multicolumn{3}{c|}{\cellcolor{gray!0}{StanfordCars}}  & \multicolumn{3}{c|}{\cellcolor{gray!0}{Flowers102}} & \multicolumn{3}{c|}{\cellcolor{gray!0}{Food101}} & \multicolumn{3}{c}{\cellcolor{gray!0}{FGVCAircraft}}    \\
    \cline{2-13} 
    &  \multicolumn{1}{c}{Base} & \multicolumn{1}{c}{New} & \multicolumn{1}{c|}{H} &  \multicolumn{1}{c}{Base} & \multicolumn{1}{c}{New} & \multicolumn{1}{c|}{H}  &  \multicolumn{1}{c}{Base} & \multicolumn{1}{c}{New} & \multicolumn{1}{c|}{H}  &  \multicolumn{1}{c}{Base} & \multicolumn{1}{c}{New} & \multicolumn{1}{c}{H}  \\
    \hline
    {CoOp} \cite{zhou2022learning} & 76.88 & 66.01 & 71.03 & 97.63 & 66.17 & 78.88 & 88.67 & 86.13 & 87.38 & 38.74 & 27.07 & 31.87 \\
    \textbf{+SAR} & 76.69 & \textbf{72.18} & \textbf{74.37} & 97.44 & \textbf{75.06} & \textbf{84.80} & \textbf{90.14} & \textbf{91.01} & \textbf{90.57} & \textbf{39.16} & \textbf{33.37} & \textbf{36.03} \\
    \hline
    {KgCoOp} \cite{yao2023kgcoop} & 73.70 & 74.04 & 73.87 & 95.66 & 73.07 & 82.85 & 90.62 & 91.48 & 91.05 & 37.48 & 32.17 & 34.62 \\
    \textbf{+SAR} & 72.69 & \textbf{75.46} & \textbf{74.05} & 95.38 & \textbf{75.32} & \textbf{84.17} & 90.59 & \textbf{91.86} & \textbf{91.22} & 36.69 & \textbf{33.51} & \textbf{35.03} \\
    \hline
    {TCP} \cite{yao2024tcp} & 79.94 & 74.03 & 76.87 & 97.88 & 74.85 & 84.83 & 90.69 & 91.37 & 91.03 & 42.02 & 34.37 & 37.81 \\
    \textbf{+SAR} & 79.59 & \textbf{74.49} & \textbf{76.96} & 97.85 & \textbf{74.97} & \textbf{84.90} & \textbf{90.72} & \textbf{91.63} & \textbf{91.17} & 41.48 & \textbf{34.97} & \textbf{37.95} \\
    \hline
    {MaPLe} \cite{khattak2023maple} & 72.51 & 74.30 & 73.39 & 96.20 & 75.32 & 84.49 & 90.67 & 92.04 & 91.35 & 35.15 & 32.41 & 33.72 \\
    \textbf{+SAR} & \textbf{72.65} & \textbf{74.63} & \textbf{73.63} & 95.66 & 74.78 & 83.94 & \textbf{90.70} & \textbf{92.05} & \textbf{91.37} & \textbf{37.15} & \textbf{37.35} & \textbf{37.25} \\
    \hline
    {CoPrompt} \cite{roy2024coprompt} & 73.16 & 70.49 & 71.80 & 96.90 & 75.13 & 84.64 & 90.17 & 91.75 & 90.95 & 35.91 & 30.31 & 32.87 \\
    \textbf{+SAR} & 72.80 & \textbf{70.98} & \textbf{71.88} & 96.77 & \textbf{75.67} & \textbf{84.93} & \textbf{90.39} & 91.73 & \textbf{91.06} & \textbf{36.67} & \textbf{37.07} & \textbf{36.87} \\
\hline
     \multicolumn{1}{l|}{\multirow{2}{*}{\makecell[c]{{Method}}}}  & \multicolumn{3}{c|}{\cellcolor{gray!0}{SUN397}}  & \multicolumn{3}{c|}{\cellcolor{gray!0}{DTD}} & \multicolumn{3}{c|}{\cellcolor{gray!0}{EuroSAT}}  & \multicolumn{3}{c}{\cellcolor{gray!0}{UCF101}}  \\
    \cline{2-13} 
    &  \multicolumn{1}{c}{Base} & \multicolumn{1}{c}{New} & \multicolumn{1}{c|}{H} &  \multicolumn{1}{c}{Base} & \multicolumn{1}{c}{New} & \multicolumn{1}{c|}{H}  &  \multicolumn{1}{c}{Base} & \multicolumn{1}{c}{New} & \multicolumn{1}{c|}{H}  &  \multicolumn{1}{c}{Base} & \multicolumn{1}{c}{New} & \multicolumn{1}{c}{H}   \\
     \hline
{CoOp} \cite{zhou2022learning} &80.89 &67.93 &73.85 &78.74 &45.65 &57.79 &91.23 &55.25 &68.82 &84.97 &63.06 &72.39\\
\textbf{+SAR} &\textbf{81.63} &\textbf{73.88} &\textbf{77.56} &\textbf{80.75} &\textbf{59.38} &\textbf{68.44} &90.48 &\textbf{69.01} &\textbf{78.30} &\textbf{85.28} &\textbf{76.22} &\textbf{80.50}\\
\hline
{KgCoOp} \cite{yao2023kgcoop} &80.67 &75.93 &78.23 &80.05 &54.31 &64.71 &87.92 &67.95 &76.66 &83.68 &74.20 &78.66\\
\textbf{+SAR} &80.48 &\textbf{77.27} &\textbf{78.84} &79.63 &\textbf{56.28} &\textbf{65.95} &87.66 &66.61 &75.70 &\textbf{83.80} &\textbf{76.44} &\textbf{79.95}\\
\hline
{TCP} \cite{yao2024tcp} &82.64 &78.12 &80.32 &82.68 &57.04 &67.51 &89.95 &75.03 &81.82 &87.21 &81.02 &84.00\\
\textbf{+SAR} &82.52 &\textbf{78.26} &\textbf{80.33} &82.56 &\textbf{58.82} &\textbf{68.70} &\textbf{89.96} &74.38 &81.43 &\textbf{87.32} &\textbf{81.48} &\textbf{84.30}\\
\hline
{MaPLe} \cite{khattak2023maple}&80.77 &77.99 &79.36 &80.63 &58.89 &68.07 &90.85 &67.95 &77.75 &83.59 &77.83 &80.61\\
\textbf{+SAR} &\textbf{81.01} &\textbf{79.00} &\textbf{79.99} &\textbf{80.75} &\textbf{63.12} &\textbf{70.85} &\textbf{92.67} &\textbf{73.52} &\textbf{81.99} &\textbf{84.06} &\textbf{79.97} &\textbf{81.96}\\
\hline
{CoPrompt} \cite{roy2024coprompt} &82.23 &79.46 &80.82 &82.72 &62.48 &71.19 &93.29 &69.57 &79.70 &86.18 &78.87 &82.36\\
\textbf{+SAR} &82.08 &\textbf{79.70} &\textbf{80.87} &82.64 &\textbf{64.53} &\textbf{72.47} &92.16 &\textbf{79.53} &\textbf{85.38} &86.12 &\textbf{79.99} &\textbf{82.94}\\
\hline
    \end{tabular}
    \vspace{+0.0em}
    \caption{Performance comparison of five baselines in base-to-new generalization w/ or w/o applying SAR on 11 datasets.}
    \label{tab:b2n}
\end{table*}

\begin{figure*}[!t]
    \centering
    \includegraphics[width=0.92 \linewidth]{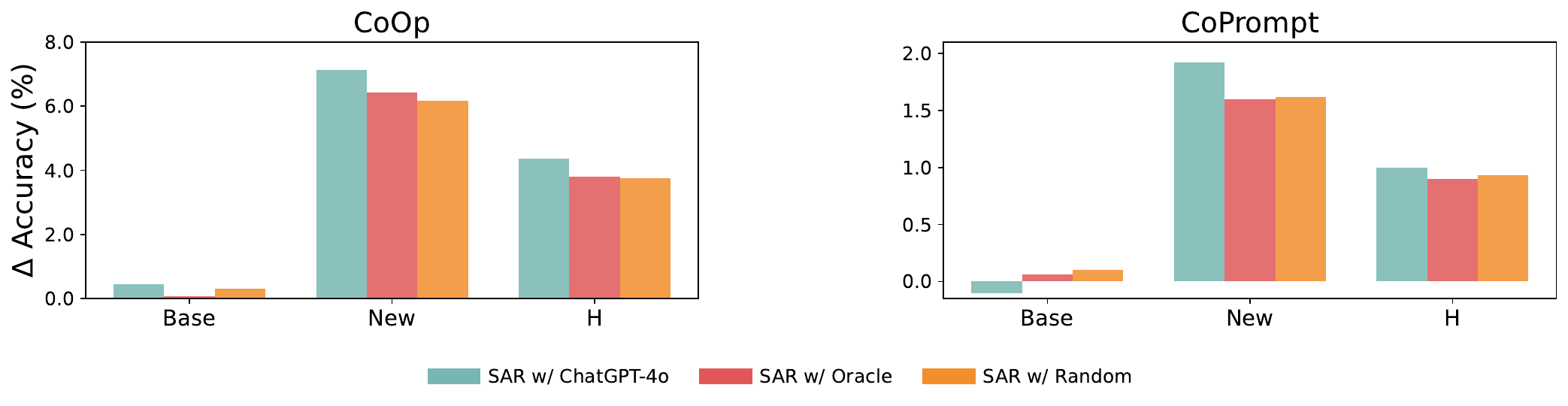}
    \caption{Performance gains of SAR over the baseline across different word sources. `Oracle' refers to the use of new classes in the dataset as novel classes for SAR. The results are averaged across 11 datasets.}
    \label{fig:wordsource_ab}
\end{figure*}

\subsection{Base-to-New Generalization}
In the Base-to-New generalization task, we evaluate the performance of models with and without SAR in 16-shot setting across 11 datasets. The classes in each dataset are divided into two groups, where prompts trained on one group (base classes) are used to evaluate performance on the other group (new classes). As summarized in \Cref{tab:b2n}, SAR consistently improves the new class accuracy and the harmonic mean (H) across all baselines. For CoOp and MaPLe, SAR also delivers significant improvements in base class accuracy, while showing no critical base-new accuracy trade-offs for other baselines. Notably, SAR yields the largest improvements for CoOp, while also bringing meaningful improvements for the baselines equipped with regularization techniques designed to improve generalization. These results suggest that SAR effectively addresses gaps that cannot be resolved by existing regularization methods of the baselines. SAR brings marginal improvements over TCP, which can be attributed to the fact that TCP, by generating class-aware prompts, already captures the semantics of new classes effectively (as shown in \Cref{fig:motivation}), thereby reducing the necessity for SAR's contribution. Similarly, the limited improvements by SAR on the Caltech101 and OxfordPets datasets are likely because all baselines already generalize well on these datasets. This claim is supported by the observation that all baselines exhibit fairly high new accuracy on these datasets, as shown in \Cref{tab:b2n}, leaving little room for SAR to provide further improvement.

\subsection{Effectiveness with Alternative Word Sources}
We investigate whether SAR remains effective when using novel classes extracted from sources other than LLMs. To this end, we train SAR with novel classes from (1) 200 randomly sampled nouns from WordNet \cite{miller1995wordnet}, a large lexical database, and (2) new class names from the dataset. The base-to-new generalization results for CoOp and CoPrompt are presented in \Cref{fig:wordsource_ab}. A key observation is that new accuracy improves significantly even when randomly sampled words are used as novel classes. This suggests that SAR is robust to the choice of novel classes and consistently enhances generalization to unseen classes. Notably, using words generated by ChatGPT-4o as novel classes leads to even greater performance improvements. This can be attributed to the fact that these words are more semantically aligned with the new classes in the dataset than randomly sampled words, making them more effective at capturing the semantics of new classes (Examples of generated words are listed in \underline{\textbf{Supp. Material C}}). However, using the actual new class names from the dataset as novel classes (SAR w/ `Oracle') does not yield the highest performance improvements. This is primarily due to the small number of them (on average, less than 100), which reduces the semantic diversity available for SAR and ultimately makes it less effective to capture the fine-grained semantics of them.

\subsection{Effectiveness in Few Shot Setting}
To validate the effectiveness of SAR in a few-shot setting, we conduct base-to-new generalization experiments under 4-shot setting for CoOp, TCP, and CoPrompt. In the case of CoOp, the models show a stronger tendency to increase SAR loss during prompt tuning compared to the 16-shot setting. This may be due to the model attempting to capture spurious correlations in the limited training data, leading to overfitting. To address this issue, we set the regularization weight to 1.0 for CoOp across all datasets. For TCP and CoPrompt, the same regularization weights as those used in the 16-shot setting are applied to each dataset. The performance averaged across 11 datasets is summarized in \Cref{tab:4shot_b2n}, demonstrating that SAR remains consistently effective even in the few-shot setting.

\begin{table}[!t]
    \small \centering
    \renewcommand{\arraystretch}{0.9}
    \setlength{\tabcolsep}{8pt}
    \scalebox{0.8}[0.8]{
    \begin{tabular}{l cc | c }
    \toprule
    Method  & Base & New & H \\
    \midrule
    CoOp & 78.11 & 66.99 & 72.12 \\
    \cellcolor{gray!20}{\textbf{+SAR}} 
    & \cellcolor{gray!20}\textbf{78.73} \textcolor{blue}{(+0.62\%)} 
    & \cellcolor{gray!20}\textbf{75.00} \textcolor{blue}{(+8.01\%)} 
    & \cellcolor{gray!20}\textbf{76.82} \textcolor{blue}{(+4.70\%)} \\
    \midrule
    TCP & \textbf{79.98} & 74.59 & 77.19 \\
    \cellcolor{gray!20}{\textbf{+SAR}} 
    & \cellcolor{gray!20}79.93 \textcolor{red}{(-0.05\%)} 
    & \cellcolor{gray!20}\textbf{75.04} \textcolor{blue}{(+0.45\%)} 
    & \cellcolor{gray!20}\textbf{77.41} \textcolor{blue}{(+0.22\%)} \\
    \midrule
    CoPrompt & 78.44 & 75.57 & 76.98 \\
    \cellcolor{gray!20}{\textbf{+SAR}} 
    & \cellcolor{gray!20}\textbf{78.49} \textcolor{blue}{(+0.05\%)} 
    & \cellcolor{gray!20}\textbf{76.59} \textcolor{blue}{(+1.02\%)} 
    & \cellcolor{gray!20}\textbf{77.53} \textcolor{blue}{(+0.55\%)} \\
    \bottomrule
    \end{tabular}
    }
    \vspace{+0.0em}
    \caption{Performance comparison of three baselines in the 4-shot setting w/ and w/o applying SAR. The results are averaged across 11 datasets.}
    \label{tab:4shot_b2n}
    \vspace{+0.0em}
\end{table}

\begin{table}[!t]
    \small \centering
    \renewcommand{\arraystretch}{1.1}
 \setlength{\tabcolsep}{8pt}
    \scalebox{0.75}[0.75]{
    \begin{tabular}{l cccccc}
    \toprule
    & \textbf{Source} & \multicolumn{5}{c}{\textbf{Target}} \\ \cmidrule(lr){2-2} \cmidrule(lr){3-7}
     & ImageNet & Avg. & -S & -R & -A  & -V2 \\
    \midrule
        CoOp \cite{zhou2022learning} &  71.49 & {59.36} & 47.85  & 75.57  & 49.71  & {64.32} \\
    \cellcolor{gray!20}{\textbf{+SAR}} & \cellcolor{gray!20}{\textbf{71.60}} & \cellcolor{gray!20}{\textbf{60.02}} & \textbf{48.55} & \textbf{76.72} & \textbf{50.38} & \textbf{64.42}  \\
    \midrule
        KgCoOp \cite{yao2023kgcoop} & \textbf{70.56}  & 59.74 & 48.54  & 76.58 & \textbf{50.27} & \textbf{63.57}  \\ 
        \cellcolor{gray!20}{\textbf{+SAR}} & \cellcolor{gray!20}{70.44} & \cellcolor{gray!20}\textbf{59.75} & \textbf{48.67} & \textbf{76.66} & 50.19 & 63.46  \\
    \midrule
        TCP \cite{yao2024tcp} & \textbf{72.53} & 59.61 & 48.11  & 76.10 & 49.43 & {64.78}  \\ 
        \cellcolor{gray!20}{\textbf{+SAR}} & \cellcolor{gray!20}{\textbf{72.53}} & \cellcolor{gray!20}{\textbf{59.74}} & \textbf{48.85} & \textbf{76.25} & \textbf{49.59} & \textbf{64.85}  \\
    \midrule
        MaPLe \cite{khattak2023maple} & 70.40 & 60.15 & 49.10  & 77.05 & 50.52 & {63.94}  \\
        \cellcolor{gray!20}{\textbf{+SAR}} & \cellcolor{gray!20}{\textbf{70.56}} & \cellcolor{gray!20}{\textbf{60.20}} & 49.02 & 77.03 & \textbf{50.67} & \textbf{64.07}  \\
    \midrule
        CoPrompt \cite{roy2024coprompt} & 71.00 & 60.27 & 50.12  & 77.82 & 48.37 & {64.77}  \\
        \cellcolor{gray!20}{\textbf{+SAR}} & \cellcolor{gray!20}{\textbf{71.01}} & \cellcolor{gray!20}\textbf{60.28} & 50.09 & 77.76 & \textbf{48.56} & 64.71  \\
    \bottomrule
    \end{tabular}}\vspace{+0.0em}
        \caption{Performance comparisons of five baselines in domain generalization w/ or w/o applying SAR.} 
    \label{tab:domain_gen}
    \vspace{-0em}
\end{table}

\subsection{Domain Generalization}
Domain generalization evaluates how well a model can generalize to domains with data distributions that differ from its training domain. Following convention, we use ImageNet as the source domain and evaluate the performance of the trained model on ImageNet-Sketch, ImageNet-A, ImageNet-R, and ImageNet-V2. The results, summarized in \Cref{tab:domain_gen}, show that SAR consistently improves the average target accuracy across the baselines.

\section{Conclusion and Limitation}
In this paper, we found that prompts trained for base classes can disrupt the semantics of unseen classes, generating text embeddings with incorrect semantic relationships among classes. To address this issue, we proposed SAR, a method that regularizes learnable prompts to preserve the similarity relationships among classes generated by hand-crafted prompts. Our method utilizes ChatGPT-4o to generate novel classes that are semantically aligned with the base classes and uses them as potential unseen classes during prompt tuning. Extensive experiments across five baselines and 11 datasets demonstrate the effectiveness of SAR in improving generalization to unseen classes. Despite its effectiveness, SAR incurs additional memory and training time to compute text embeddings for novel classes. As future work, we aim to reduce the resource cost of SAR while exploring performance enhancement strategies that do not rely on hand-crafted prompts as supervison.

{
    \small
    \bibliographystyle{ieeenat_fullname}
    \bibliography{main}
}

\clearpage
\appendix
\setcounter{page}{1}
\maketitlesupplementary

\definecolor{codegreen}{rgb}{0,0.6,0}
\definecolor{codegray}{rgb}{0.5,0.5,0.5}
\definecolor{codepurple}{rgb}{0.58,0,0.82}
\definecolor{backcolour}{rgb}{0.95,0.95,0.92}

\section{Prompt for ChatGPT-4o}

\begin{lstlisting}[
    language={},
    label=lst:gpt,
    backgroundcolor=\color{backcolour},   
    commentstyle=\color{codegreen},
    keywordstyle=\color{magenta},
    numberstyle=\tiny\color{codegray},
    stringstyle=\color{codepurple},
    basicstyle=\ttfamily\footnotesize,
    breakatwhitespace=false,         
    breaklines=true,                 
    keepspaces=true,                 
    numbers=none,        
    numbersep=5pt,                  
    showspaces=false,                
    showstringspaces=false,
    showtabs=false,                  
    tabsize=2,
    frame=single,
    rulecolor=\color{black}
]
You are a helpful assistant that identifies the common, specific concept among a given set of words and provides additional words that fit within that concept. For example, given ["rose," "daisy," "tulip"], you might infer the concept of "plants," with a more specific focus on "flowers." Once you've identified the concept, generate exactly 200 words that belong to it.

The new words should be as semantically distinct from each other as possible while staying relevant to the shared concept. Aim for diversity within the category to showcase a broad range of examples. Please return the words in a Python list format.

Here are the my words: [`Annual Crop Land', 'Forest', 'Herbaceous Vegetation Land', 'Highway or Road', 'Industrial Buildings']
\end{lstlisting}

We provide the prompt given to ChatGPT-4o for generating novel classes for EuroSAT. The prompt is designed based on the templates from \cite{anperceptionclip}. For other datasets, only the list of base classes is modified.

\section{Prompts for Ensembling}
\begin{table}[ht]
    \centering
    \renewcommand{\arraystretch}{1.0}
    \label{tab:pe}
    \resizebox{\columnwidth}{!}{
    \begin{tabular}{p{2.8cm} p{9cm}}
    \toprule
    \textbf{Dataset} & \textbf{Hand-crafted prompt} \\
    \midrule
    \multirow{7}{*}{All} & ``\texttt{itap of a [CLASS].}'' \\ 
                         & ``\texttt{a bad photo of the [CLASS].}'' \\ 
                         & ``\texttt{a origami [CLASS].}'' \\ 
                         & ``\texttt{a photo of the large [CLASS].}'' \\ 
                         & ``\texttt{a [CLASS] in a video game.}'' \\ 
                         & ``\texttt{art of the [CLASS].}'' \\ 
                         & ``\texttt{a photo of the small [CLASS].}'' \\ 
    \midrule
    ImageNet & ``\texttt{a photo of a [CLASS].}'' \\
    Caltech101 & ``\texttt{a photo of a [CLASS].}'' \\ 
    OxfordPets & ``\texttt{a photo of a [CLASS], a type of pet.}'' \\
    StanfordCars & ``\texttt{a photo of a [CLASS].}'' \\
    Flowers102 & ``\texttt{a photo of a [CLASS], a type of flower.}'' \\
    Food101 & ``\texttt{a photo of [CLASS], a type of food.}'' \\
    FGVCAircraft & ``\texttt{a photo of an aircraft [CLASS].}'' \\
    SUN397  & ``\texttt{a photo of a [CLASS].}'' \\
    DTD & ``\texttt{a photo of a [CLASS], a type of texture.}'' \\
    EuroSAT & ``\texttt{a centered satellite photo of [CLASS].}'' \\
    UCF101 & ``\texttt{a photo of a person doing [CLASS].}'' \\
    \bottomrule
    \end{tabular}}
\end{table}

We use the prompt templates provided in the implementation of CoOp \cite{zhou2022learning} for prompt ensembling. For each dataset, a total of eight prompts are ensembled, comprising seven general prompts and one dataset-specific custom prompt.

\section{Word Examples}
We compare the words generated by ChatGPT-4o with the actual new classes in the dataset. This qualitative comparison suggests that the words generated by ChatGPT-4o are semantically aligned with the new classes, making them suitable candidates for unseen classes.

\begin{table}[ht]
    \centering
    \footnotesize
    \setlength{\tabcolsep}{4pt}
    \renewcommand{\arraystretch}{0.90}
    \label{tab:wordex}

    \begin{subtable}{\linewidth}
        \centering
        \subcaption{DTD}
        \vspace{2mm}
        \begin{tabular}{p{1.5cm} >{\raggedright\arraybackslash}p{6cm}}
        \toprule
        & \textbf{Words} \\
        \midrule
        Base class & \texttt{banded, blotchy, braided, bubbly, bumpy, chequered, cobwebbed, cracked, crosshatched, crystalline, ...} \\
        \midrule
        ChatGPT-4o & \texttt{acid-washed, aerated, airy, angular, anodized, antiqued, arced, asymmetrical, beaded, bizarre, ...} \\
        \midrule
        New class & \texttt{matted, meshed, paisley, perforated, pitted, pleated, polka-dotted, porous, potholed, scaly, ...} \\
        \bottomrule
        \end{tabular}
    \end{subtable}

    \vspace{3mm}

    \begin{subtable}{\linewidth}
        \centering
        \subcaption{FGVCAircraft}
        \vspace{2mm}
        \begin{tabular}{p{1.5cm} >{\raggedright\arraybackslash}p{6cm}}
        \toprule
        & \textbf{Words} \\
        \midrule
        Base class & \texttt{707-320, 727-200, 737-200, 737-300, 737-400, 737-500, 737-600, 737-700, 737-800, 737-900, ...} \\
        \midrule
        ChatGPT-4o & \texttt{Airbus A220-100, Airbus A220-300, Airbus A300, Airbus A310, Airbus A320neo, Airbus A321neo, ...} \\
        \midrule
        New class & \texttt{Cessna 560, Challenger 600, DC-10, DC-3, DC-6, DC-8, DC-9-30, DH-82, DHC-1, DHC-6, ...} \\
        \bottomrule
        \end{tabular}
    \end{subtable}

    \vspace{3mm}

    \begin{subtable}{\linewidth}
        \centering
        \subcaption{StanfordCars}
        \vspace{2mm}
        \begin{tabular}{p{1.5cm} >{\raggedright\arraybackslash}p{6cm}}
        \toprule
        & \textbf{Words} \\
        \midrule
        Base class & \texttt{2000 AM General Hummer SUV, 2012 Acura RL Sedan, 2012 Acura TL Sedan, 2008 Acura TL Type-S, ...} \\
        \midrule
        ChatGPT-4o & \texttt{2019 Ford Mustang Coupe, 2020 Toyota Camry Sedan, 2018 Honda Civic Hatchback, 2021 Chevrolet Tahoe SUV, ...} \\
        \midrule
        New class & \texttt{2012 FIAT 500 Abarth, 2012 FIAT 500 Convertible, 2012 Ferrari FF Coupe, 2012 Ferrari California Convertible, ...} \\
        \bottomrule
        \end{tabular}
    \end{subtable}

\end{table}

\end{document}